\documentclass[11pt]{article}
\usepackage[preprint]{acl}
\usepackage{times}
\usepackage[most]{tcolorbox}
\usepackage{pgfplots}\pgfplotsset{compat=1.18}
\usepackage[export]{adjustbox}
\usepackage{subcaption}
\usepackage{siunitx}
\usepackage{latexsym}
\usepackage[T1]{fontenc}
\usepackage[utf8]{inputenc}
\usepackage{microtype}
\usepackage{inconsolata}
\usepackage{enumitem}
\usepackage{graphicx}
\usepackage{amsmath}
\usepackage{booktabs}
\usetikzlibrary{arrows.meta, positioning, calc}
\usepackage{fontawesome5}
\usepackage{multirow}
\tcbset{
  card/.style={enhanced, colback=gray!3, colframe=black!55, drop shadow={black!25},
    coltitle=white, colbacktitle=black!62, fonttitle=\small, boxrule=0.6pt, arc=3pt,
    left=6pt, right=6pt, top=4pt, bottom=6pt},
  srcA/.style={enhanced, colback=blue!3, colframe=blue!55!black,
    coltitle=white, colbacktitle=blue!55!black, fonttitle=\scriptsize,
    boxrule=0.5pt, arc=2pt, left=3pt, right=3pt, top=3pt, bottom=3pt},
  srcB/.style={enhanced, colback=red!3, colframe=red!55!black,
    coltitle=white, colbacktitle=red!55!black, fonttitle=\scriptsize,
    boxrule=0.5pt, arc=2pt, left=3pt, right=3pt, top=3pt, bottom=3pt},
}
\usepackage{placeins}  % provides \FloatBarrier to keep tables in their section
\usepackage{url}       % \url for the anonymized artifact link

\title{Which Source Wins? Task-Dependent Reliance in Vision--Language Models}

\author{
{\normalfont Rodela Ghosh}\thanks{Equal contribution.} \\
University of South Florida \\
\texttt{rg21@usf.edu}
\And
{\normalfont Aviral Gupta}\footnotemark[1] \\
University of South Florida \\
\texttt{aviralgupta@usf.edu}
\And
{\normalfont Guangjing Wang} \\
University of South Florida \\
\texttt{guangjingwang@usf.edu}
}

\begin{document}

\maketitle

\begin{abstract}
Vision-language models (VLMs) combine images and text, but when the two conflict and one becomes harder to read, it is unclear how a model shifts its reliance between them. We study this \emph{modality reallocation} with a controlled setup: we degrade either the image or the text across four levels of legibility while keeping the other clean, and track how the model's preference changes. We build conflicts from GSM8K and SVAMP by pairing the rendered image of one arithmetic problem with the text of another, so the two sources support different answers. We also introduce ChartQA-Conflict, a manually reviewed benchmark of 229 chart-report conflicts with matched chart and table-image representations. We evaluate six open-weight VLMs using both generated answers and a length-normalized conditional log-likelihood margin. On GSM8K and SVAMP, five of six models shift more strongly away from degraded text than from degraded images. On ChartQA-Conflict, all six likelihood-scored models exhibit the opposite pattern, shifting more strongly away from the degraded visual source. This reversal persists after calibrating for unimodal accuracy loss and after replacing charts with plain table images. Two frontier API models, GPT-5.6-Luna and Gemini-3.5-Flash, behaviorally replicate the ChartQA-Conflict reversal, with GPT-5.6-Luna also matching the arithmetic direction. These results show that modality reliance in VLMs is not fixed, but varies across tasks, evidence structures, models, and evaluation settings. The source code is available at \url{https://github.com/Ro-netizen004/multimodal-arbitration-artifact}.
\end{abstract}

\section{Introduction}
Humans integrate multisensory cues and down-weight noisier sources when cues conflict~\cite{ernst2002optimal,ernst2004merging}. 
Vision-language models (VLMs) likewise integrate visual and textual inputs, yet it remains unclear whether they appropriately recalibrate their reliance on each modality when one becomes less reliable.

\begin{figure}[t]
\centering
\begin{tcolorbox}[card, width=\linewidth,
  title={\faRobot~\textbf{Model input}}]
\small
\ttfamily You are given two sources describing a math problem.
Source A is the attached image. Source B is the text below. Solve the problem step by step and end with \#\#\#\# <answer>.
\normalfont

\vspace{5pt}
% ---- Source A: the attached image ----
\begin{tcolorbox}[srcA, title={\faImage~Source A\, — attached image}]
  \centering\includegraphics[width=0.92\linewidth]{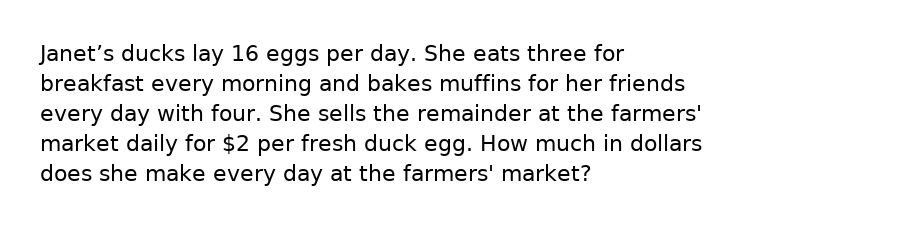}
\end{tcolorbox}

\vspace{3pt}
% ---- Source B: the text ----
\begin{tcolorbox}[srcB, title={\faAlignLeft~Source B\, — text}]
  \ttfamily A robe takes 2 bolts of blue fiber and half that much white
  fiber. How many bolts in total does it take?
\end{tcolorbox}
\normalfont
\end{tcolorbox}
\caption{The VLM is prompted to solve a problem based on two independent information sources. 
\textbf{Source~A} is an image, which renders one problem; \textbf{Source~B} is the text of a \emph{different} problem.}
\label{fig:arithmetic_example}
\vspace{-10pt}
\end{figure}

Prior work shows that VLMs perform worse when text-based problems are shown in an image rather than provided as plain text~\cite{yuan2025gsm8kv,xu2026crossmath,liu2026vista}. This performance gap has been attributed in part to the additional challenge of extracting information from images before reasoning over the recovered content~\cite{sun2026readingnotthinking}. However, existing evaluations typically assess visual and textual inputs in isolation, which cannot determine which modality a model relies on when both are presented but provide conflicting evidence.
In this paper, we ask: when one source becomes harder to read, does a VLM shift its reliance toward the other, and does it shift by the same amount whether we degrade the image or the text? An asymmetric shift would indicate that modality reweighting depends not only on degradation levels but also on which modality is degraded and how the model is evaluated in a specific task.

To answer the question, we build multiple datasets for evaluation. (i) We create controlled image-text conflicts using GSM8K~\citep{cobbe2021gsm8k} and SVAMP~\citep{patel2021nlp} arithmetic datasets. Specifically, we render an arithmetic problem from plain text into a picture. Then, we pair the rendered image of one arithmetic problem with the text of another, so the two sources support different answers, as shown in Figure~\ref{fig:arithmetic_example}. We define four legibility levels and then degrade either the image or the text into different legibility levels while keeping the other source clean. To avoid implying a primary modality, we refer to the inputs as Source~A and Source~B and counterbalance these labels across items.
(ii) We further construct ChartQA-Conflict from ChartQA~\cite{masry2022chartqa}, pairing each chart with a textual report that supports a conflicting answer. This dataset tests whether the observed modality-reliance patterns generalize beyond rendered-text images.

We evaluate six open-weight VLMs in controlled image-text conflicts across three benchmarks. The experiments reveal a task-dependent asymmetry: on the arithmetic benchmarks (GSM8K and SVAMP), five of six models shift more strongly away from degraded text than from degraded images. By contrast, on ChartQA-Conflict, all six likelihood-scored models shift more strongly away from the degraded visual source. This reversal persists after calibrating for unimodal accuracy loss and after replacing charts with plain table images, indicating that VLMs do not exhibit a fixed preference for text or vision. Two frontier API models, GPT-5.6-Luna and Gemini-3.5-Flash, behaviorally replicate the ChartQA-Conflict reversal, indicating the pattern is not confined to open-weight models. Instead, modality reallocation depends on the task, evidence structure, model, and, to a lesser extent, prompt framing.

In summary, our contributions are:

\begin{itemize}
\item \textbf{A new problem formulation and conflict-based evaluation datasets.}
We introduce the problem of \emph{modality reallocation}: how VLMs redistribute their reliance between visual and textual sources when the two conflict and one becomes harder to read. We construct controlled conflicts from GSM8K and SVAMP and introduce ChartQA-Conflict, a manually reviewed benchmark of 229 chart-report conflicts with matched chart and table-image representations.

\item \textbf{A controlled framework for measuring source reliance.}
We degrade either the visual or textual source across four legibility levels while keeping the other source clean, counterbalance source labels, and quantify preference using both generated answers and a length-normalized conditional log-likelihood margin. This design enables within-item comparisons of how strongly models shift away from each degraded modality.

\item \textbf{Evidence that modality reallocation is setting-dependent rather than fixed.}
Across arithmetic conflicts, five of six models shift more strongly away from degraded text, whereas all six CLL-scored models show the opposite pattern on ChartQA-Conflict. This reversal persists after calibration for unimodal accuracy loss and after replacing charts with table images, and is behaviorally replicated by two frontier API models (GPT-5.6-Luna and Gemini-3.5-Flash), showing that VLM reliance depends on the task, evidence structure, and model rather than reflecting a universal preference for text or vision.
\end{itemize}

\section{Related Work}

\paragraph{VLM modality reliance preference.}
VLMs often underperform when problems are presented as images rather than native text~\cite{yuan2025gsm8kv,xu2026crossmath,liu2026vista}. However, these studies compare performance across input formats and do not examine which modality a model follows when visual and textual sources provide conflicting answers. 
More broadly, VLMs are documented to under-use visual information and default to textual cues~\cite{jain2025wordspixels}.
However, \citet{hua2025conflicting} pair images with contradictory captions and show, at a fixed conflict level, that source preference is systematic, model-dependent, and reflected in internal representations. \citet{ortu-etal-2026-seeing} studies conflicts between visual evidence and a model's stored knowledge using logit-based analysis and causal interventions. SimpleOCR~\citep{peng-etal-2026-simpleocr} also renders questions as images to study the modality-utilization gap. Yet, SimpleOCR uses rendering to improve visual reading rather than to induce cross-modal conflict.
Our study extends this line of work in three ways. First, we examine conflicts in multi-step reasoning tasks rather than in simple object-recognition tasks. Second, we make either the image or the text harder to read across four legibility levels and track how the model's preference changes for the same item. Third, we use counterbalanced Source~A/B labels so that neither source is framed as the primary input.

\paragraph{Properties of conflicting inputs.}
Prior studies vary the properties of conflicting evidence. Some degrade the textual information~\cite{deng2025blindfaith}. Others change reasoning difficulty or uncertainty~\cite{pezeshkpour2025mixed,zhang2025uncertainty}. A further line asks whether visual evidence is internally represented even when the answer follows the text~\cite{nooralahzadeh2026arbitration}, and whether chain-of-thought traces can look visually grounded while in fact following the text~\cite{villegas2026reasoning}. Each of these manipulates one source. We instead degrade the image and the text in item-matched degradation arms with nominally aligned levels, and compare how source preference changes in each. We do not assume the two degradation levels are equally severe, and we design a secondary analysis that adjusts for severity using the measured drop in unimodal accuracy. CMC-Bench~\citep{catapang-2026-image} creates image-text conflicts using ChartQA and MMMU~\cite{yue2024mmmu}, but varies the conflict type to evaluate accuracy and abstention~\citep{moratelli-etal-2026-benchmarking}. In contrast, we keep each conflict fixed, degrade each source separately across four levels, and measure within-item shifts in source preference using generated answers and length-normalized likelihood margins.

\begin{figure*}[t]
\centering
\resizebox{0.95\textwidth}{!}{%
\begin{tikzpicture}[
  font=\small, >={Latex[length=2.2mm]},
  box/.style={rounded corners=3pt, draw, line width=0.6pt, align=center,
              inner sep=3.5pt, minimum width=30mm, minimum height=10mm},
  vis/.style={box, fill=green!8,  draw=green!55!black},
  txt/.style={box, fill=red!7,    draw=red!60!black},
  oc/.style={box, fill=violet!8,  draw=violet!60!black},
  arr/.style={->, draw=black!45, line width=0.6pt},
  hdr/.style={font=\bfseries},
]
\node[hdr] at (0,1.15)  {Conflict construction};
\node[hdr] at (6,1.15)  {Degrade one source};
\node[hdr] at (12,1.15) {Outcome measures};
\node[vis] (vis) at (0,0)    {\bfseries\textcolor{green!45!black}{Visual source} $a_I$\\[1pt]{\scriptsize\textcolor{black!55}{rendered problem or chart}}};
\node[txt] (txt) at (0,-1.5) {\bfseries\textcolor{red!55!black}{Textual source} $a_T$\\[1pt]{\scriptsize\textcolor{black!55}{problem text or report}}};
\node[vis] (dvis) at (6,0)    {\bfseries\textcolor{green!45!black}{Degrade visual}\\[1pt]{\scriptsize\textcolor{black!55}{textual stays clean $\cdot$ 4 levels}}};
\node[txt] (dtxt) at (6,-1.5) {\bfseries\textcolor{red!55!black}{Degrade textual}\\[1pt]{\scriptsize\textcolor{black!55}{visual stays clean $\cdot$ 4 levels}}};
\node[oc] (gen) at (12,0)    {\bfseries\textcolor{violet!60!black}{Generated choice}\\[1pt]{\scriptsize\textcolor{black!55}{which answer it produces}}};
\node[oc] (cll) at (12,-1.5) {\bfseries\textcolor{violet!60!black}{CLL margin}\\[1pt]{\scriptsize\textcolor{black!55}{length-normalized log-likelihood}}};
\foreach \a in {vis,txt}{\foreach \b in {dvis,dtxt}{\draw[arr] (\a) -- (\b);}}
\foreach \a in {dvis,dtxt}{\foreach \b in {gen,cll}{\draw[arr] (\a) -- (\b);}}
\node[draw=black!45, dashed, rounded corners=3pt, align=center, inner sep=3.5pt,
      minimum width=112mm] at (6,-2.75)
      {\bfseries Role-neutral prompt\\[1pt]{\scriptsize\textcolor{black!55}{visual and textual sources labeled Source A\,/\,B, counterbalanced across items}}};
\node[font=\scriptsize] at (6,-3.7)
  {Two settings:\quad image\,+\,text problem (GSM8K, SVAMP)\quad$\cdot$\quad chart\,+\,report (ChartQA-Conflict)};
\end{tikzpicture}}
\caption{Experimental logic. We pair the rendered image of one problem with the text of another so the two sources support different answers ($a_I$ vs.\ $a_T$). A role-neutral prompt labels them Source~A/B, counterbalanced across items. From the clean pair we degrade one source at a time across four legibility levels while the other stays clean, and score every trial by both the generated source choice and the length-normalized CLL preference margin.}
\label{fig:overview}
\end{figure*}
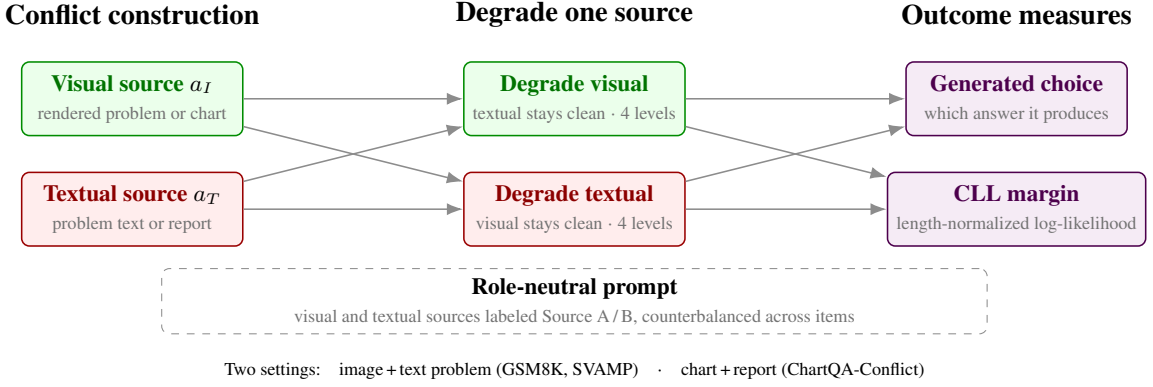

\section{Methodology}
In this section, we evaluate how a model redistributes its reliance between the image and the text when one of them becomes harder to read, which is referred to as \emph{source reallocation}.

The pipeline is the same across two conflict settings as shown in Figure~\ref{fig:overview}. In both settings, the model receives an image and text that support different answers. We degrade one source across four legibility levels while keeping the other clean. We then measure source preference using both the generated answer and a length-normalized CLL margin (\S\ref{sec:legibility_method}). For each model, we compare whether preference shifts more strongly when the image is degraded or when the text is degraded.

The two settings differ in how the conflict is constructed. In \emph{rendered-text conflict} on GSM8K and SVAMP, the image and text contain different arithmetic problems in natural language. In \emph{natural-visual conflict} on ChartQA-Conflict, both sources address the same question, but the chart and the accompanying report support different answers. The same degradation, measurement, and comparison procedures (\S\ref{sec:legibility_method}) apply to both settings.

\subsection{Models}
We evaluate six open-weight VLMs spanning four families and 2--8B parameters: Qwen2-VL-2B~\citep{wang2024qwen2vl}, Qwen2.5-VL-7B~\citep{bai2025qwen25vl}, LLaVA-1.6-Mistral-7B~\citep{liu2024improved,liu2024llavanext}, LLaVA-OneVision-7B~\citep{li2024llavaonevision}, Idefics3-8B~\citep{laurencon2024idefics3}, and Phi-3.5-Vision-4B~\citep{abdin2024phi3}. All six models are evaluated using conditional log-likelihood. Generated-answer evaluation is also conducted for all six, although Phi-3.5-Vision yields too few decidable ChartQA-Conflict responses for reliable behavioral analysis. A seventh model, InternVL2-8B~\citep{chen2024internvl2}, and two frontier API models, GPT-5.6-Luna and Gemini-3.5-Flash, lack the continuation-scoring interface the CLL margin requires and are used for behavioral (generated-answer) evidence only, reported separately (\S\ref{sec:frontier_results}, Appendix~\ref{app:frontier}). Models are loaded without quantization and decoded using \texttt{bfloat16} where supported. Full software versions are in Appendix~\ref{app:implementation}.

\subsection{Dataset Construction}
\subsubsection{Rendered image conflict}
\label{sec:rendered}

We pair the rendered image of one natural language-based math problem with the native text of a different problem. 
Since the two problems support different answers, the model's response
indicates which source it followed. 
Specifically, image problem $i$ is paired with text problem $(i+1)\bmod N$, where $N$ is the number of benchmark problems.
Equivalently, text problem $i$ is paired with image problem $(i-1)\bmod N$. Figure~\ref{fig:arithmetic_example} shows a worked example, where a response of $a_I{=}18$ is image-following and $a_T{=}3$ is text-following. For odd-indexed items, the Source~A/B labels are swapped, so neither modality is always Source~A.

We exclude pairs whose two problems have the same answer to avoid confusing the follow-up preference analysis. Images are rendered as 900-pixel-wide PNGs using DejaVu Sans, black text on a white background. Each image contains the same information as its native-text version, so only the input format differs. We also evaluate each problem separately in text-only and image-only conditions.

\subsubsection{Natural visual conflict}
\label{sec:chartqa}

\paragraph{Construction.}
Each ChartQA-Conflict item pairs a native chart supporting answer $a_I$ with an
accompanying textual report giving counterfactual evidence for a different answer
$a_T$ to the same question (Figure~\ref{fig:chartqa_example}); unlike the
rendered-image setting (\S\ref{sec:rendered}), the chart carries native visual
structure (bars, axes, legends). We build items from the official ChartQA test set and
source tables~\citep{masry2022chartqa}, choosing each $a_T$ to preserve the original
answer's type and unit and stay semantically plausible. Two of the authors, both
senior undergraduates, independently reviewed all 230 items, confirming that the
report supports $a_T$, that $a_T$ is a valid answer, and that $a_T \neq a_I$ after
normalization; a post-run audit removed one item whose report did not support its
answer, leaving 229 conflicts (excluded identically across models, arms, and levels).
Full details are in Appendix~\ref{app:chartqa_construction}.

\paragraph{Prompt.}
ChartQA-Conflict uses a role-neutral prompt: the shared question, a statement
that the model is given ``two conflicting evidence sources'' with ``neither
source privileged,'' then the chart and report under counterbalanced Source A/B
labels (A first). Template in Appendix~I.

\paragraph{Chart-versus-table control.}
To test whether the reversal depends on the chart's graphical encoding rather
than on any image-format visual, we replace the chart with a plain-table image of
the same official source table, holding the question, report, conflict, and
degradation ladder fixed. Both representations present identical facts, so both
still support $a_I$; only the visual form changes. Details in Appendix~I.

\begin{figure*}[t]
\centering
\setlength{\fboxsep}{1pt}
\begin{tcolorbox}[card, width=\textwidth,
  title={\textbf{Legibility degradation ladders (same conflict item)}}]

% ---------- Image arm ----------
\begin{tabular}{@{}c@{\hspace{6pt}}cccc@{}}
\multirow{2}{*}{\rotatebox{90}{\small\textbf{\textcolor{blue!55!black}{Image}}}} &
\includegraphics[width=0.205\textwidth, cframe=blue!55!black]{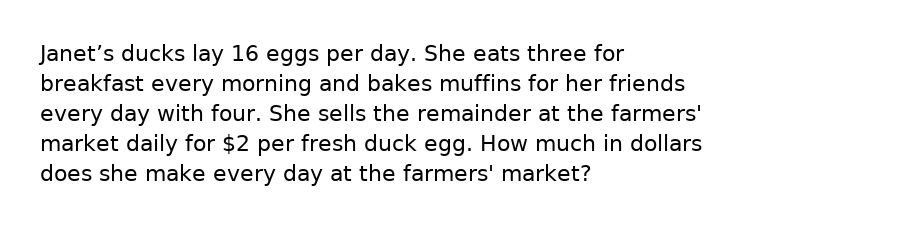} &
\includegraphics[width=0.205\textwidth, cframe=blue!55!black]{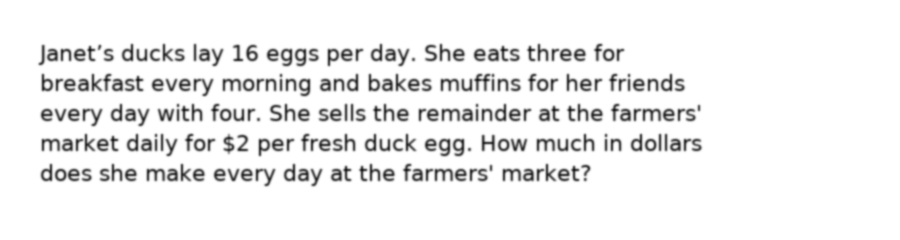} &
\includegraphics[width=0.205\textwidth, cframe=blue!55!black]{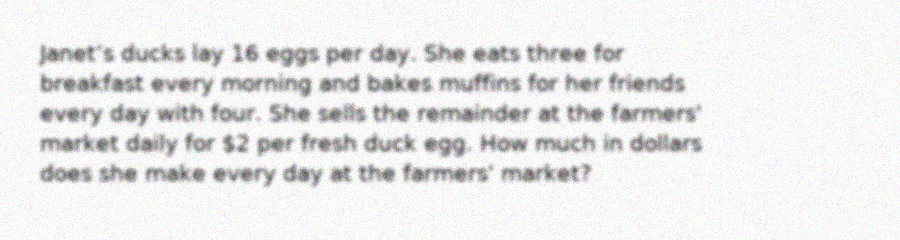} &
\includegraphics[width=0.205\textwidth, cframe=blue!55!black]{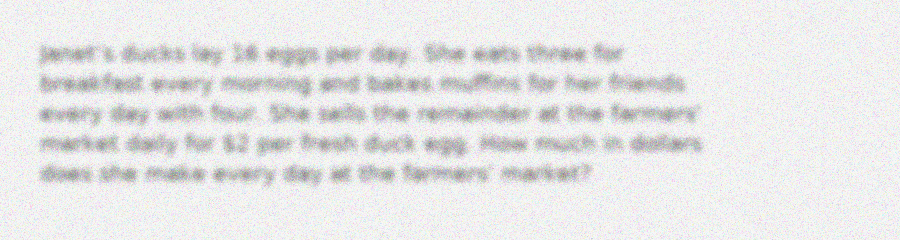} \\[1pt]
 & {\footnotesize L0 $\cdot$ clean} & {\footnotesize L2 $\cdot$ light blur} &
   {\footnotesize L4 $\cdot$ blur+noise} & {\footnotesize L5 $\cdot$ heavy} \\
\end{tabular}

\vspace{6pt}

% ---------- Text arm ----------
\begin{tabular}{@{}c@{\hspace{6pt}}p{0.205\textwidth}p{0.205\textwidth}p{0.205\textwidth}p{0.205\textwidth}@{}}
\multirow{2}{*}{\rotatebox{90}{\small\textbf{\textcolor{red!55!black}{Text}}}} &
\fcolorbox{red!55!black}{red!4}{\parbox[t]{0.18\textwidth}{\scriptsize\ttfamily A robe takes 2 bolts of blue fiber and half that much white fiber. How many bolts in total does it take?}} &
\fcolorbox{red!55!black}{red!4}{\parbox[t]{0.18\textwidth}{\scriptsize\ttfamily A robe take 2 bUlts o lue fiber and half that much white fiber. How many bolts in total doe6 it take?}} &
\fcolorbox{red!55!black}{red!4}{\parbox[t]{0.18\textwidth}{\scriptsize\ttfamily robe ta1es 2 boltb f bluC fiber and haf that much white fiber. How m1y bTlts ia total does it ake?}} &
\fcolorbox{red!55!black}{red!4}{\parbox[t]{0.18\textwidth}{\scriptsize\ttfamily rCbe 5akeU 2 bbts bf blu fiber ad alf tha4 much white fib1. HTw mana bols in oal does i tke}} \\[1pt]
 & {\footnotesize L0 $\cdot$ clean} & {\footnotesize L2 $\cdot$ light} &
   {\footnotesize L4 $\cdot$ medium} & {\footnotesize L5 $\cdot$ heavy} \\
\end{tabular}
\end{tcolorbox}
\caption{Legibility ladders for the same conflict item. \textbf{Top (image arm):}
the rendered image (Source~A) is degraded from clean (L0) to heavy blur$+$noise (L5)
with the text clean. \textbf{Bottom (text arm):} the conflicting text (Source~B) is
corrupted at the character level with the image clean. L0/L2/L4/L5 mark increasing
corruption within each arm, matched by level index, not absolute severity.}
\label{fig:degradation_ladders}
\end{figure*}

\subsection{Matched Degradation and Measurement}
\label{sec:legibility_method}

We test whether models shift away from a source as that source becomes harder to read. Each experiment therefore has two parallel degradation arms. In the image-degradation arm, the image is degraded while the text remains clean. In the text-degradation arm, the text is degraded while the image remains clean. Across arms and degradation levels, we keep the examples, source-supported answers, models, decoding procedure, and scoring method fixed.

\paragraph{Legibility levels.}
We evaluate four prespecified levels: clean (L0), light degradation (L2), moderate degradation (L4), and heavy degradation (L5). These labels come from a larger image-corruption ladder~\citep{hendrycks2019robustness} used earlier in the project. We omit L1 and L3 because they contain JPEG-only corruptions with no direct text counterpart. Thus, L0/L2/L4/L5 provide a shared clean-to-heavy progression for both modalities.

\paragraph{Degradation.}
We evaluate four aligned degradation levels: clean (L0), light (L2), moderate (L4), and heavy (L5). Images are degraded using increasing Gaussian blur and pixel noise, while text is degraded by randomly deleting or replacing non-whitespace characters, with whitespace preserved to keep word boundaries visible. Figure~\ref{fig:degradation_ladders} illustrates both degradation ladders for one conflict item.
The same image operations apply to the ChartQA
charts. All corruption is deterministic and seeded. Every model receives identical
degraded inputs and the same corrupted text is used for the generation and CLL
scoring. The exact blur radii, noise levels, and corruption rates per-level are in
Appendix~\ref{app:degradation}.

\paragraph{Behavioral source choice.}
At each level, we classify the generated response as image-following,
text-following, neither, ambiguous, or invalid. A trial is \emph{decidable} when
its extracted answer matches exactly one source-supported answer. Neither,
ambiguous, and invalid responses are retained and reported but excluded from
source-preference estimates; we also report the decidable count because
degradation can affect both source choice and answer attributability.

We first extract the answer following the requested \texttt{\#\#\#\#} delimiter.
For ChartQA-Conflict, some models instead return a bare answer on the first line.
When the delimiter is absent, we accept the first nonempty line
if that entire line normalizes to a single valid answer. We do not search
explanations or reasoning traces for embedded numbers. For GSM8K and SVAMP, the
extracted numeric answer is rounded to the nearest integer. For
ChartQA-Conflict, we normalize superficial formatting differences, including
commas, signs, decimal notation, percentages, currency symbols, compatible
units, and yes/no variants, and then require an exact match to one candidate.
Equal normalized candidates or responses matching both are ambiguous, and
responses with no valid extracted answer are invalid. We use no fuzzy matching.

\paragraph{Continuous source-preference margin.}
Generated choice can be insensitive when the model already prefers one source at
the clean level L0. This is common in the image arm, where several models start
out favoring text even as their underlying preference keeps shifting. We
therefore add a continuous measure from the probabilities of the two candidate
answers. For a conflict trial, let $a_T$ and $a_I$ be the text- and
image-supported answers, $c$ the shared multimodal prompt, and $|a_T|,|a_I|$
their token lengths:
\begin{equation}
m(c) = \frac{1}{|a_T|}\log p(a_T\mid c) - \frac{1}{|a_I|}\log p(a_I\mid c).
\label{eq:margin}
\end{equation}
Each $\log p(a\mid c)$ sums the teacher-forced log-probabilities of the candidate
tokens, giving a mean log-probability per answer token (nats/token); a positive
margin favors the text answer, a negative one the image answer. Length
normalization reduces but does not remove the advantage of shorter
answers~\citep{holtzman2021surface}; Appendix~\ref{app:cll_normalization} shows
that across GSM8K, SVAMP, and ChartQA-Conflict the exponent rescales the
asymmetry's magnitude but preserves its sign and significance for every model.
Only answer tokens after the fixed \texttt{\#\#\#\#} delimiter are scored, with
no chain-of-thought, so the margin measures direct-answer preference, not
reasoning-trace attribution.
As a validity check, $\operatorname{sign}(m)$ agrees with the generated source choice
on 75.4\% of 26{,}893 decidable model--item--level observations (six models, GSM8K and
SVAMP, both degradation arms).
More detailed analysis is in Appendix~\ref{app:cll_validation}.
Since the two agree and the margin is graded and available
for all six models, we adopt it as the primary measure for the reallocation analysis and
use generated choice as a behavioral cross-check.

\paragraph{Comparing the two arms.}
Let $m^{(I)}_{i,L}$ and $m^{(T)}_{i,L}$ denote the margins for item $i$ at level $L$
in the image- and text-degradation arms. We orient both changes toward the source
that remains clean:
\[
R_{I,i} = m^{(I)}_{i,L5}-m^{(I)}_{i,L0}, \qquad
R_{T,i} = m^{(T)}_{i,L0}-m^{(T)}_{i,L5}.
\]
So $R_{I,i}>0$ means item $i$ moves toward the text-supported answer as the image
degrades, and $R_{T,i}>0$ means it moves toward the image-supported answer as the text
degrades. The within-item arm asymmetry is
\[
A_i = R_{T,i}-R_{I,i},
\]
with $A_i>0$ indicating stronger reallocation under text degradation and $A_i<0$
stronger reallocation under image degradation. For each model we report the medians of
$R_{I,i}$, $R_{T,i}$, and $A_i$; because the median of $A_i$ is computed within item,
it need not equal the difference of the two separately reported medians. Statistical
testing is described in \S\ref{sec:stats}.

\subsection{Benchmarks}
\label{sec:benchmarks}

\paragraph{Matched-degradation benchmarks.}
The primary rendered-arithmetic experiments use the complete GSM8K test
set~\citep{cobbe2021gsm8k} (1{,}319 problems) and a fixed 300-item SVAMP
subset~\citep{patel2021nlp}. In each benchmark, image item $i$ is paired with text item
$(i+1)\bmod N$, with wraparound at the end, and this pairing is fixed across models,
arms, levels, generation, and CLL scoring. The two paired answers are distinct under
the numerical matching rule for 1{,}304 GSM8K pairs and 296 SVAMP pairs; equal-answer
pairs are retained in the saved outputs but marked ambiguous and excluded from the
decidable-preference denominator. The same pairings are used in both degradation arms.

ChartQA-Conflict provides a separate natural-visual evaluation, using an original
chart and an evidence-bearing report that support different answers to the same
question. After excluding one item that failed the post-run entailment audit, the
analysis contains 229 valid conflicts; its construction and attribution are described
in \S\ref{sec:chartqa}.

\paragraph{Data release.}
We release our annotations, degradation metadata, conflict mappings, and
generated/CLL outputs, together with the reviewed ChartQA-Conflict dataset and
exclusion record
(\url{https://huggingface.co/datasets/vlm-modality-research/chartqa-evidence-conflict-v2}),
the chart/table ablation set
(\url{https://huggingface.co/datasets/vlm-modality-research/chartqa-evidence-conflict-table-v1}),
and the rendered GSM8K/SVAMP conflict stimuli
(\url{https://huggingface.co/datasets/vlm-modality-research/modality-conflict-arbitration-v2}).
Upstream licenses are unchanged (GSM8K/SVAMP: MIT; ChartQA: GPL-3.0).

\subsection{Statistical Analysis}
\label{sec:stats}

\paragraph{Within- and between-arm changes.}
CLL results are matched by item ID. For each item and arm we compute the L0-to-L5 margin
change, summarize the item-level changes by their median, and test whether the paired differences are centered at zero with a two-sided Wilcoxon signed-rank test~\citep{wilcoxon1945individual}, a 95\% paired-bootstrap CI (10{,}000 resamples)~\citep{efron1993bootstrap},
and a paired randomization test (10{,}000 draws; fixed seed). The primary cross-arm
comparison is the within-item asymmetry $A_i$ (\S\ref{sec:legibility_method}), tested
with the same procedures; $A_i>0$ means stronger reallocation under text degradation.

\paragraph{Adjustment for unimodal task loss.}
Image and text degradation at the same nominal level may not remove the same
amount of usable information. We therefore estimate degradation severity using
each model's accuracy when it receives only the source being degraded. For model
$M$, channel $c$, and level $L$, we calculate the proportional accuracy loss
relative to the clean condition, $\ell_{M,c,L}$. We then fit, separately for
each model and degradation arm, a line relating this loss to the median shift
in preference toward the source that remains clean. The slope $b_{M,c}$
therefore measures reallocation per unit of lost single-modality accuracy. We then compare the text- and image-degradation slopes:
\[
D_M = b_{M,T} - b_{M,I}.
\]
A positive $D_M$ indicates stronger reallocation when text is degraded than when the
image is degraded, after accounting for the measured accuracy loss. Confidence
intervals are obtained by bootstrapping matched items. We treat this as a secondary
analysis because each slope is estimated from only three degraded levels and
single-modality accuracy is itself an imperfect estimate of source usability.
Additional pooled-regression and character/OCR-survival analyses
\citep{smith2007tesseract} are reported in
Appendix~\ref{app:legibility_adjustment}.

\section{Results}
\label{sec:results}

\begin{figure}[t]
  \centering
  \includegraphics[width=0.49\textwidth]{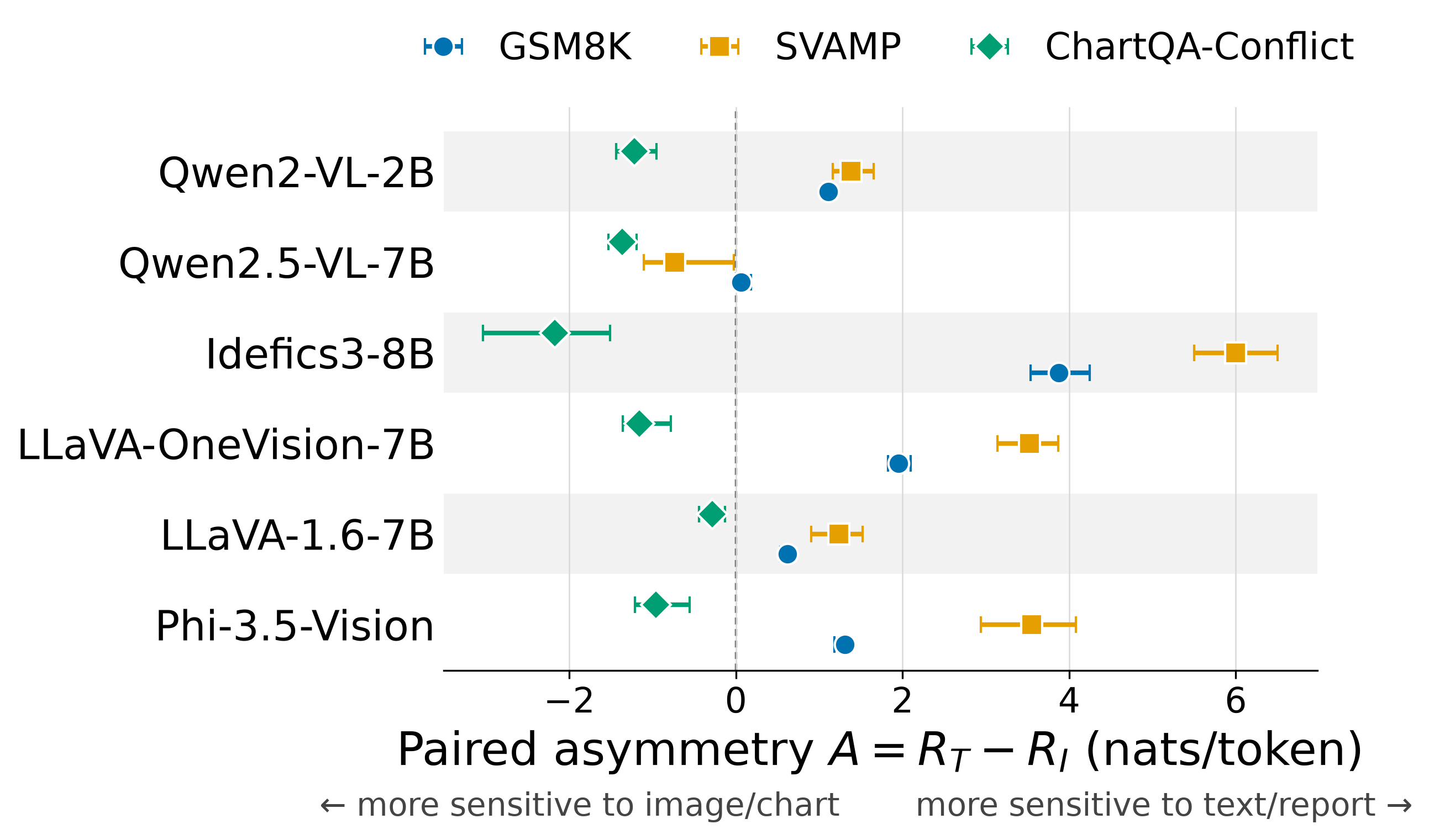}
  \caption{Paired asymmetry $A=R_T-R_I$ per model and setting; whiskers are 95\%
  paired-bootstrap CIs. $A>0$ indicates a stronger response to text/report
  degradation, $A<0$ to image/chart degradation. On GSM8K and SVAMP the
  asymmetry is positive for five of six models; on ChartQA-Conflict it reverses
  to negative for all six. Full per-model values are in
  Table~\ref{asymmetry_combined}.}
  \label{fig:asymmetry_forest}
\end{figure}

Figure~\ref{fig:asymmetry_forest} summarizes the findings across different settings: on the arithmetic conflicts (GSM8K, SVAMP), the paired asymmetry is positive
for five of six models, whereas on ChartQA-Conflict it reverses to negative for all six. Table~\ref{asymmetry_combined} reports the full per-model $R_I$, $R_T$, and paired asymmetry $A$ with 95\% bootstrap confidence intervals. We first evaluate the consistency of two measures (\S\ref{sec:measure_agreement}), then examine the arithmetic
pattern (\S\ref{sec:main_asymmetry}) and the reversal (\S\ref{sec:chartqa_results}).

\begin{table*}[t]
\centering
\footnotesize
\setlength{\tabcolsep}{6pt}
\begin{tabular}{llrrrc}
\toprule
Benchmark & Model
& \(R_I\)
& \(R_T\)
& \(A\) (paired)
& Paired-bootstrap 95\% CI for \(A\) \\
\midrule
\multirow{6}{*}{GSM8K}
& Qwen2-VL-2B        & \(+0.063\) & \(+1.193\) & \(+1.107\) & \([+1.031,+1.191]\) \\
& Qwen2.5-VL-7B      & \(+0.855\) & \(+1.013\) & \(+0.059\) & \([+0.000,+0.181]\) \\
& Idefics3-8B        & \(+0.750\) & \(+4.844\) & \(+3.877\) & \([+3.540,+4.250]\) \\
& LLaVA-OneVision-7B & \(+0.022\) & \(+1.982\) & \(+1.950\) & \([+1.829,+2.099]\) \\
& LLaVA-1.6-7B       & \(+0.006\) & \(+0.645\) & \(+0.616\) & \([+0.531,+0.670]\) \\
& Phi-3.5-Vision     & \(+0.064\) & \(+1.403\) & \(+1.310\) & \([+1.184,+1.390]\) \\
\midrule
\multirow{6}{*}{SVAMP}
& Qwen2-VL-2B        & \(+0.502\) & \(+1.995\) & \(+1.385\) & \([+1.165,+1.650]\) \\
& Qwen2.5-VL-7B      & \(+3.079\) & \(+2.597\) & \(-0.731\) & \([-1.103,-0.022]\) \\
& Idefics3-8B        & \(+1.781\) & \(+7.634\) & \(+6.000\) & \([+5.500,+6.500]\) \\
& LLaVA-OneVision-7B & \(+0.081\) & \(+3.482\) & \(+3.524\) & \([+3.139,+3.869]\) \\
& LLaVA-1.6-7B       & \(+0.194\) & \(+1.353\) & \(+1.234\) & \([+0.904,+1.519]\) \\
& Phi-3.5-Vision     & \(+0.016\) & \(+3.515\) & \(+3.548\) & \([+2.940,+4.083]\) \\
\midrule
\multirow{6}{*}{\shortstack[l]{ChartQA-\\Conflict}}
& Qwen2-VL-2B        & \(+2.955\) & \(+1.686\) & \(-1.221\) & \([-1.440,-0.953]\) \\
& Qwen2.5-VL-7B      & \(+3.242\) & \(+1.779\) & \(-1.362\) & \([-1.528,-1.190]\) \\
& Idefics3-8B        & \(+7.375\) & \(+4.483\) & \(-2.175\) & \([-3.033,-1.507]\) \\
& LLaVA-OneVision-7B & \(+2.748\) & \(+1.514\) & \(-1.156\) & \([-1.359,-0.780]\) \\
& LLaVA-1.6-7B       & \(+1.979\) & \(+1.589\) & \(-0.280\) & \([-0.443,-0.133]\) \\
& Phi-3.5-Vision     & \(+3.452\) & \(+2.561\) & \(-0.963\) & \([-1.210,-0.558]\) \\
\bottomrule
\end{tabular}
\caption{\textbf{Reliance on the clean source when the other is degraded (neutral
prompt).} \(R_I\)/\(R_T\) is the median reallocation toward the clean text/image
when the image/text is degraded (chart/report on ChartQA-Conflict); \(A\) is the
median within-item contrast \(R_{T,i}-R_{I,i}\), with \(A>0\) (\(A<0\)) indicating
stronger reallocation under text (image) degradation. As medians are not additive,
\(A\) need not equal \(R_T-R_I\). Intervals are paired-bootstrap 95\% CIs
(10{,}000 resamples; \(n=229\) for ChartQA-Conflict).}
\label{asymmetry_combined}
\end{table*}

\subsection{Behavioral Choice and CLL Consistency}
\label{sec:measure_agreement}

The two measures are moderately consistent. Across six models, the CLL margin favors the
same source as the generated answer on $75.4\%$ of $26{,}893$ decidable trials
($95\%$ CI: $74.4$--$76.4\%$; Appendix~\ref{app:cll_validation}), and both shift in the same direction as degradation
increases. We therefore base the reallocation analysis on the CLL margin---graded and
available for all six models---and use generated answers as a behavioral cross-check,
since generated choices can saturate near the ceiling.

\subsection{Text Degradation Produces Stronger Reallocation on Arithmetic Conflicts}
\label{sec:main_asymmetry}

Recall that a positive item-level contrast
\(A_i=R_{T,i}-R_{I,i}\) means reallocation is stronger under text than image
degradation. On GSM8K the median asymmetry is positive for all six models, and for
five the paired-bootstrap interval excludes zero (median \(A_i\) up to \(+3.877\)
nats/token); Qwen2.5-VL-7B is the exception, with a near-zero estimate
(\(A=+0.059\)) whose interval reaches zero. The same five models show the pattern
on SVAMP, while Qwen2.5-VL-7B reverses (\(A=-0.731\), 95\% CI \([-1.103,-0.022]\)).
The arithmetic result is therefore a consistent five-of-six cross-model pattern,
not a universal property of every model.

\subsection{Calibration by Unimodal Task Loss}
\label{sec:calibrated_slopes_results}

Calibrating each arm's reallocation by the model's proportional loss of
unimodal accuracy reproduces the same task-dependent pattern
(Appendix~\ref{app:legibility_adjustment},
Tables~\ref{tab:arithmetic_calibrated_slopes}
and~\ref{tab:chartqa_calibrated_slopes}). On GSM8K and SVAMP, five of six
models have a positive slope difference \(D=b_T-b_I\) with bootstrap intervals
excluding zero; Qwen2.5-VL-7B is again the arithmetic exception, with \(D<0\)
on both. On ChartQA-Conflict the direction reverses: all six models
have \(D<0\) with intervals excluding zero.
The reversal survives adjustment for degradation severity, supporting
task- and format-dependence rather than a fixed modality preference. We treat
this as secondary, since each slope uses only three degraded levels and
conditions on the accuracy estimates.

\subsection{Prompt Framing Modulates Effect Size}
\label{sec:prompt_framing_results}

The asymmetry remains positive across five models with matched GSM8K CLL
results (Appendix~\ref{app:prompt_framing}, Table~\ref{tab:prompt_framing}) under
both the original and role-neutral prompts, showing that framing alone does not
create the arithmetic asymmetry. Neutral framing changes its magnitude for four models
(\(p<10^{-3}\)) but not Phi-3.5-Vision (\(p=.112\)), with effects varying in
direction. The largest change is for Qwen2.5-VL-7B, whose median asymmetry falls
from \(+0.813\) to \(+0.060\). Thus, framing modulates the effect in a
model-dependent way without explaining the broader pattern.

\subsection{The Asymmetry Reverses for Natural Visual Evidence}
\label{sec:chartqa_results}

Unlike the arithmetic conflicts, where both sources are essentially text (one just
shown as an image), ChartQA-Conflict pits a real chart against a written report,
and this flips the result (Figure~\ref{fig:asymmetry_forest}, ChartQA series;
per-model values in Table~\ref{asymmetry_combined}). Across the 229 items all six
models show a negative asymmetry (\(-0.280\) to \(-2.175\) nats/token) with every
confidence interval below zero: each moves away from the chart more strongly when
the chart is degraded than away from the report when the report is degraded. This
reverses the arithmetic pattern for all six models and persists after adjusting for
how much each degradation lowers accuracy
(\S\ref{sec:calibrated_slopes_results}).

The reversal also holds behaviorally: generated answers are negative for all five
decidable open models (InternVL2 the exception; Phi-3.5-Vision has too few
decidable items), and both frontier models replicate it (\S\ref{sec:frontier_results}). We treat the six-model CLL results as primary and these behavioral
results as converging support
(Appendices~\ref{app:chartqa_behavior} and~\ref{app:frontier}).

\paragraph{Graphical encoding alone does not explain the reversal.}
To see whether the reversal comes from the chart's graphical form (its bars, axes,
and plotted marks), we replace each chart with a table image showing the same
numbers, keeping the question, report, conflict, and degradation levels unchanged.
All six models still show a negative asymmetry with the table image
(Table~\ref{tab:chart_table}, Appendix~\ref{app:chart_table}), so the chart's
graphical marks are not necessary for the reversal. This does not identify the
cause on its own, since the table image still differs from rendered text in
layout and the effect is inconsistent across models, but it shows the reversal
survives removing the chart's graphical encoding.

\subsection{Evaluation on Frontier Larger Models}
\label{sec:frontier_results}

We also evaluate generated-answer behavior from GPT-5.6-Luna and Gemini-3.5-Flash.
As these APIs expose no teacher-forced scores, we use exact answer attribution only
and treat it as supporting evidence, separate from the primary six-model CLL analysis.
On the 229 ChartQA-Conflict items both models reproduce the natural-visual direction
(negative paired asymmetry, \(p<.001\)); on 300 role-neutral GSM8K conflicts
GPT-5.6-Luna instead matches the open-model arithmetic direction (positive,
\(p<.001\)). Full statistics, trajectories, and the clean-endpoint robustness check are detailed in
Appendix~\ref{app:frontier}.

\section{Conclusion}

We introduced a controlled framework for studying \emph{modality reallocation} in VLMs by presenting conflicting visual and textual evidence, degrading each source in turn, and measuring source preference using both generated answers and conditional log-likelihood margins. Across GSM8K, SVAMP, and the newly constructed ChartQA-Conflict benchmark, we find that VLMs do not reallocate reliance symmetrically. Five of six models shift more strongly away from degraded text on arithmetic conflicts, whereas all six CLL-scored models shift more strongly away from degraded visual evidence on ChartQA-Conflict. This reversal persists after calibration for unimodal accuracy loss and under the chart-to-table control, showing that modality reliance is not a fixed bias toward text or vision but a setting-dependent behavior shaped by the task, evidence structure, model, and prompt framing. By providing a new problem formulation, curated conflict datasets, and a reproducible evaluation methodology, this work offers the research community a systematic basis for diagnosing how multimodal models arbitrate between competing sources.

\section*{Limitations}

First, the current conflict construction is designed for tasks with attributable numeric answers. It does not transfer directly to multiple-choice benchmarks such as AQuA-RAT~\citep{ling2017aqua}, where models may output only an option label, and the selected source cannot always be identified reliably. Extending the framework to free-form, categorical, and multiple-choice outputs will require alternative attribution procedures.

Second, our model coverage is limited to open-weight VLMs with 2B--8B parameters because the CLL analysis requires access to token-level log-probabilities. Closed frontier models are therefore excluded from the primary graded analysis. In addition, the prompt-framing comparison covers only five models on GSM8K, and the legibility-adjusted regression contains six model clusters. The results should thus be interpreted as evidence for the evaluated models rather than as a model-family-wide generalization.

Third, the two conflict settings do not isolate task, representation, and conflict construction independently. In GSM8K and SVAMP, the visual source is rendered text, whereas ChartQA-Conflict uses a chart and report that answer the same question. The chart-versus-table control shows that chart-specific graphical encoding alone does not explain the reversal, but the remaining differences between arithmetic and ChartQA-Conflict prevent us from identifying a single causal mechanism.

Fourth, ChartQA-Conflict is limited to one natural-visual benchmark and was reviewed
by two of the authors, who were not blind to the counterfactual design, rather than
independent annotators. Moreover, the answer-relevant value appears in the report as
a stated figure in the text, listed alongside values for other categories or years,
so recovering it requires reading and matching the text to the question but not visual
chart-reading; obtaining the corresponding value from a chart or table may
additionally require visual localization and interpretation. The reversal therefore
reflects reallocation between sources with potentially different baseline directness,
not a comparison between equally accessible visual and textual evidence.

Fifth, the visual and textual corruption ladders are matched by nominal level rather than psychometric severity. We partially address this issue by calibrating reallocation against the accuracy loss produced when each degraded source is presented alone. However, unimodal accuracy is an approximate measure of source usability, and the calibrated slopes are estimated from three degraded levels.

Finally, our framework measures forced source arbitration under reduced legibility rather than factual reliability or conflict awareness. Degradation makes a source harder to read without making its content less correct, and the prompt requires a single answer. Consequently, we do not test whether models detect the contradiction, express uncertainty, or abstain. The CLL margin also agrees with generated source choices on 75.4\% of attributable trials, indicating that it is a useful but imperfect proxy for behavioral reliance.

\section*{Ethical Considerations}

This work presents limited direct ethical risk, but its findings should not be interpreted as evidence that any evaluated model is safe or reliable for deployment. Our experiments examine source reliance under controlled visual and textual degradation. Gaussian blur and character corruption are simplified proxies for reduced legibility and do not capture the full range of accessibility barriers encountered in practice. The results may also not generalize to other languages, scripts, or tokenization schemes because our text corruptions are restricted to English alphanumeric characters.

\section*{Declaration of Generative AI Assistance}
The authors used generative AI coding and writing assistants, specifically Claude Code (Anthropic) and Codex (OpenAI), in a limited, supervised capacity
during this work. Their use was confined to (i) software-engineering
support---debugging experiment and analysis scripts, orchestrating cluster jobs,
and formatting figures and \LaTeX{} tables---and (ii) language editing of
author-written text, including tightening prose, improving clarity, and checking
cross-references and consistency. The research questions, experimental design,
datasets, evaluation methodology, analyses, results, and conclusions were
conceived, implemented, and verified by the authors. All AI-assisted outputs were
reviewed by the authors, who take full responsibility for the entire content of
this paper, including any remaining errors.

\section*{Acknowledgments}

We thank the CIS Lab at the University of South Florida for supporting this work through access to GPU computing resources and funding for API-based model evaluations. We are also grateful for the technical guidance and discussions that helped us develop and refine our experimental setup.

\bibliography{custom}

\appendix

\section{Implementation Details}
\label{app:implementation}

\paragraph{Models and checkpoints.}
We use the following Hugging Face checkpoints. The listed revisions are the
snapshots resolved in the experiment environment; InternVL2-8B is included in
generated-answer analyses only because its model-specific chat interface does
not support the continuation-scoring procedure used for CLL.

\begin{table*}[t]
\centering
\small
\begin{tabular}{ll}
\toprule
Checkpoint & Revision \\
\midrule
\texttt{Qwen/Qwen2-VL-2B-Instruct}                & \texttt{895c3a49bc3fa70a340399125c650a463535e71c} \\
\texttt{Qwen/Qwen2.5-VL-7B-Instruct}              & \texttt{cc594898137f460bfe9f0759e9844b3ce807cfb5} \\
\texttt{HuggingFaceM4/Idefics3-8B-Llama3}         & \texttt{fddb4ff79181e55a994674777e06cd5456ce3dc3} \\
\texttt{llava-hf/llava-onevision-qwen2-7b-ov-hf}  & \texttt{0d50680527681998e456c7b78950205bedd8a068} \\
\texttt{llava-hf/llava-v1.6-mistral-7b-hf}        & \texttt{2424fdd47412fccc66d91719126b420e9fbd7065} \\
\texttt{microsoft/Phi-3.5-vision-instruct}        & \texttt{12b77fb40b63a2c73c68243d3f767aab688a1b2a} \\
\texttt{OpenGVLab/InternVL2-8B}                   & \texttt{6fb9ad6924f69424e57fab2ab061d707688f0296} \\
\bottomrule
\end{tabular}
\caption{Hugging Face checkpoints and pinned commit revisions (InternVL2-8B is
used for behavioral analysis only).}
\label{tab:checkpoints}
\end{table*}

Models are loaded with \texttt{device\_map="auto"},
\texttt{torch\_dtype=bfloat16}, and no quantization. Qwen, LLaVA, Idefics3,
and Phi use their checkpoint-specific Hugging Face processors without manual
image-resolution overrides. InternVL2 uses its model-specific chat interface;
images are converted to RGB, resized to \(448\times448\) with bicubic
interpolation, and normalized using ImageNet mean and standard deviation.
The attention backend is left at the Transformers or checkpoint default except
for Phi-3.5-Vision, for which the loader explicitly requests
\texttt{eager} attention.

\paragraph{Prompting.}
Arithmetic prompts ask the model to solve the problem step by step and end with
\texttt{\#\#\#\# <answer>}. ChartQA-Conflict instead requests exactly one
answer-only line in that format. For Qwen2-VL, Qwen2.5-VL,
LLaVA-OneVision, and Idefics3, we render the checkpoint's chat template using
\texttt{apply\_chat\_template} with
\texttt{add\_generation\_prompt=True}. LLaVA-1.6 and Phi-3.5 use the following
fixed templates, where \texttt{\{prompt\}} denotes the complete task prompt and
line breaks are literal:

\begin{verbatim}
LLaVA-1.6:
[INST] <image>
{prompt} [/INST]

Phi-3.5:
<|user|>
<|image_1|>
{prompt}<|end|>
<|assistant|>
\end{verbatim}

For CLL scoring, the same task prompt and model-specific conversation scaffold
are used. The assistant context is then extended with the literal delimiter
\texttt{\#\#\#\# } before either candidate answer is appended.

\paragraph{Generation.}
Open-model decoding is deterministic and greedy:
\texttt{do\_sample=False}, with temperature, top-\(p\), and top-\(k\) unset.
We use \texttt{max\_new\_tokens=256} for the arithmetic experiments and
\texttt{max\_new\_tokens=128} for open-model ChartQA-Conflict. Items are
generated individually (batch size one), and generation terminates at the
model's configured stopping token or the output-token limit.

For arithmetic responses, answer extraction first uses the value following
\texttt{\#\#\#\#}; if that delimiter is absent, it falls back to the final numeric
value in the response. Source attribution compares the extracted number with
the two candidate answers after rounding. ChartQA-Conflict uses a stricter
answer-only parser: it takes the value following \texttt{\#\#\#\#}, or, when that
delimiter is absent, the first nonempty line only if the entire line normalizes
to a single valid answer. It applies typed normalization and never extracts
numbers from surrounding explanations or reasoning traces.

\paragraph{Frontier API models.}
The two frontier models are accessed through hosted APIs: GPT-5.6-Luna
(OpenAI Chat Completions, \texttt{gpt-5.6-luna}) and Gemini-3.5-Flash
(Google \texttt{generate\_content}, \texttt{gemini-3.5-flash}). Because neither
API exposes teacher-forced continuation scores, both are evaluated with
generated answers only and never enter the CLL analysis. We request answer-only
outputs in the same single-line \texttt{\#\#\#\# <answer>} format used for
open-model ChartQA-Conflict and attribute responses with the identical typed,
exact-match normalization, without the numeric reasoning-trace fallback used for
the arithmetic parser. Both models use a completion budget of \(1{,}024\) tokens
and \(\text{temperature}=0\) (automatically omitted for models that reject a
non-default temperature), with low provider-side reasoning: OpenAI reasoning
effort \texttt{none} and Gemini thinking level \texttt{minimal}. We run the
image-degradation arm first and reuse its clean (L0) generations as the
report-degradation arm's L0, so the shared clean endpoint is not resampled; on
GSM8K the two L0 endpoints were generated independently, and we report the
resulting baseline sensitivity in Appendix~\ref{app:frontier}. Conflict items
are loaded from a pinned Hugging Face dataset revision, taking the deterministic
prefix of the frozen row order when a subset is used so that conflict IDs and
corruption seeds match the open-model runs. The frontier ChartQA-Conflict
evaluation uses all \(229\) items and the GSM8K behavioral evaluation \(300\)
role-neutral conflicts (GPT-5.6-Luna only). Before each full run we verified
output plausibility on a small 5-item generation probe—requiring non-empty
responses, valid A/B attribution, and exactly one \texttt{\#\#\#\# <answer>}
line—and launched the full run only after this check passed. We persist per-item
API response metadata (finish reason and token counts) for post-hoc validation.
Because hosted APIs are not fully deterministic, all frontier results are treated
as supporting behavioral evidence rather than part of the primary analysis.

\paragraph{Conditional log-likelihood scoring.}
CLL is computed using teacher forcing. The model-specific scoring context ends
with the literal string \texttt{\#\#\#\# }, including one trailing space. Candidate
strings are appended without an additional leading space. We tokenize the
context and the context--candidate concatenation with
\texttt{add\_special\_tokens=False}, and define the candidate span as
\[
\left|\operatorname{tok}(c+a)\right|
-
\left|\operatorname{tok}(c)\right|.
\]
Only this suffix is scored: the logit at position \(p-1\) predicts the token at
position \(p\). Let \(\log p(a\mid c)\) denote the summed teacher-forced
log-probability of the candidate's answer tokens; dividing by the candidate
token count \(|a|\) gives the mean log-probability per answer token in
nats/token. The arbitration margin is then identical to Eq.~(1),
\[
m(c)
=
\frac{1}{|a_T|}\log p(a_T\mid c)
-
\frac{1}{|a_I|}\log p(a_I\mid c),
\]
so a positive margin favors the text-supported answer and a negative margin
favors the image-supported answer. Both candidates are scored under the same
multimodal context. The clean (L0) condition is the fully clean image--text pair
and is therefore identical in both degradation arms, so a single L0 CLL value is
shared by $R_I$ and $R_T$ (and counted once in the measure-agreement analysis,
Appendix~\ref{app:cll_validation}).

\paragraph{Determinism and seeds.}
Open-model generation uses greedy decoding, and all corruptions are deterministic.
The image corruption for the item \(i\) is seeded by \(42+i\); text corruption is seeded by
the corresponding textual-source index. Statistical confidence intervals and
permutation tests use \(10{,}000\) resamples with fixed seeds. The exact seeds and
model-specific offsets are provided in the analysis scripts released.

\paragraph{Environment and hardware.}
Each open-model inference job used one NVIDIA L40S GPU with 48\,GB of memory.
The final experiments required approximately \texttt{400} GPU-hours in
total, excluding preliminary smoke tests and failed runs. The experiments used
Python 3.10.20 (GCC 14.3.0), PyTorch 2.5.1~\citep{paszke2019pytorch} with
CUDA 12.1 (\texttt{torch==2.5.1+cu121}), cuDNN 9.1.0, Transformers
4.49.0~\citep{wolf2020transformers}, Torchvision 0.20.1
(\texttt{torchvision==0.20.1+cu121}), Pillow 12.2.0 and Accelerate 1.14.0.

\section{Degradation Details}
\label{app:degradation}

\paragraph{Image degradation.}
For GSM8K and SVAMP, we degrade the problem images rendered. For ChartQA-Conflict,
we apply the same operations to the original charts. L2 uses Gaussian blur with
radius 1. L4 uses Gaussian blur with radius 2 followed by Gaussian pixel noise with
standard deviation 15. L5 uses Gaussian blur with radius 3, Gaussian noise with
standard deviation 25, and a contrast factor of 0.7. L0 leaves the image unchanged
and is pixel-identical to the clean image in the conflict condition. Corruption is
deterministic, using seed $42+i$ for image item $i$, so every model receives the
same degraded image.

\paragraph{Text degradation.}
In the text-degradation arm, the image remains clean while the text is corrupted.
Each non-whitespace character is independently selected with probability $0.08$ at
L2, $0.18$ at L4, and $0.35$ at L5. A selected character is deleted with probability
$0.5$ and otherwise replaced with a random alphanumeric character. Whitespace is
preserved so that word boundaries and the overall structure remain visible. Text
corruption is deterministic, seeded by the textual source's index---the paired item
$(i+1)\bmod N$ in the arithmetic setting, or the report's own item in
ChartQA-Conflict---and the same corrupted text is used for generation and CLL
scoring.

\section{Agreement Between Generated Choice and CLL Margin}
\label{app:cll_validation}

The CLL margin provides a continuous measure of direct-answer candidate
preference, whereas generated source choice records which candidate appears in
the model's final answer. To assess how closely these measures correspond, we
compare the sign of the CLL margin with the generated choice on attributable
trials. A positive margin predicts the text-supported answer and a negative
margin predicts the image-supported answer. We exclude generations matching
neither or both candidates, invalid responses, and unavailable or exactly zero
margins.

Table~\ref{tab:cll_generation_agreement} reports agreement by model and
degradation arm, pooled across GSM8K and SVAMP. After counting the clean
endpoint shared by the two arms only once, the measures agree on
\(20{,}281/26{,}893=75.4\%\) of observations (item-clustered bootstrap 95\% CI:
\(74.4\%\)--\(76.4\%\)). Agreement varies across models and is lower overall in
the text-degradation arm. Thus, the CLL margin is behaviorally grounded but is
not interchangeable with generated choice; we use it as a continuous measure
of candidate preference and report generated behavior separately.

\section{Frontier-Model Behavioral Results}
\label{app:frontier}

We evaluate GPT-5.6-Luna and Gemini-3.5-Flash using generated answers only, because
the API interfaces used in our experiments do not expose the teacher-forced
continuation scores required for the CLL margin. Attribution is exact: formatting and
compatible unit labels are normalized, but conflicting scales, currencies, or units
are rejected. Neither model enters the primary six-model CLL analysis. Throughout,
the subscript \(I\) denotes the image channel (the chart in ChartQA-Conflict, the
rendered image in GSM8K) and \(T\) the text channel (the report or the plain-text
problem); \(R_I\) and \(R_T\) are the within-item reallocations toward the clean
channel when the other is degraded, and \(A=R_T-R_I\).

\paragraph{ChartQA-Conflict endpoints.}
Table~\ref{tab:frontier_chartqa} reports the paired L0-to-L5 contrasts. Both frontier
models have negative asymmetry, reproducing the direction of the open-model
ChartQA-Conflict CLL analysis. GPT-5.6-Luna reallocates almost entirely under chart
degradation; Gemini-3.5-Flash shows a smaller but still strongly chart-dominant
contrast.

\begin{table}[t]
\centering
\small
\setlength{\tabcolsep}{4pt}
\resizebox{\columnwidth}{!}{%
\begin{tabular}{lrrrrl}
\toprule
Model & \(n\) & \(R_I\) & \(R_T\) & \(A\) & 95\% CI \\
\midrule
GPT-5.6-Luna     & 173 & \(0.925\) & \(0.000\) & \(-0.925\) & \([-0.971,-0.873]\) \\
Gemini-3.5-Flash & 171 & \(0.836\) & \(0.064\) & \(-0.772\) & \([-0.854,-0.684]\) \\
\bottomrule
\end{tabular}%
}
\caption{Generated-answer contrasts on ChartQA-Conflict. \(R_I\)/\(R_T\): within-item
reallocation toward the clean report/chart when the chart/report is degraded;
\(A=R_T-R_I\), negative \(A\) indicating stronger reallocation under chart
degradation. 95\% CIs are paired-bootstrap (10{,}000 resamples); both paired
permutation tests \(p<.001\).}
\label{tab:frontier_chartqa}
\end{table}

\paragraph{ChartQA-Conflict trajectories.}
Table~\ref{tab:frontier_chartqa_trajectories} reports preference for the source that
remains clean, among attributable generated answers at each level. Both models stay
largely chart-following under light chart degradation and move sharply toward the
report under moderate or heavy chart degradation, whereas report degradation produces
little additional movement toward the already-preferred chart.

\begin{table}[t]
\centering
\footnotesize
\setlength{\tabcolsep}{4pt}
\begin{tabular}{llrrrr}
\toprule
Model & Clean src. & L0 & L2 & L4 & L5 \\
\midrule
GPT-5.6-Luna
 & Report & 3.0  & 3.5  & 96.3 & 95.4 \\
 & Chart  & 97.0 & 98.5 & 97.5 & 97.4 \\
\midrule
Gemini-3.5-Flash
 & Report & 14.3 & 14.4 & 43.3 & 93.6 \\
 & Chart  & 85.7 & 94.3 & 95.8 & 97.3 \\
\bottomrule
\end{tabular}
\caption{Frontier-model ChartQA-Conflict trajectories: behavioral preference (\%),
among attributable generated answers, for the source that remains clean. \emph{Report}
rows track report-following as the chart is degraded; \emph{Chart} rows track
chart-following as the report is degraded. Arms coincide at L0; the report-degradation
arm reuses the image-arm L0 generations, so the L0 values are complementary.}
\label{tab:frontier_chartqa_trajectories}
\end{table}

\paragraph{GSM8K behavioral contrast.}
We also evaluate GPT-5.6-Luna on 300 role-neutral GSM8K conflicts at L0 and L5. The
model moves toward the clean source in both arms, but more strongly when the text is
degraded: \(R_I=0.323\), \(R_T=0.614\), \(A=+0.291\). The complete-case analysis
retains \(n=189\) matched items, with a paired-bootstrap 95\% CI of
\([+0.169,+0.407]\) and a two-sided paired permutation test giving \(p<.001\). This
direction agrees with the predominant open-model arithmetic result and is opposite to
GPT-5.6-Luna's ChartQA-Conflict result.

\paragraph{Clean-endpoint sensitivity.}
Unlike the frontier ChartQA experiment, which reuses the image-arm L0 generations in
the report-degradation arm, the two nominally identical GSM8K L0 endpoints were
generated independently, with behavioral text preferences of \(65.9\%\) and
\(60.2\%\) (API nondeterminism). Repeating the descriptive endpoint calculation with
each L0 sample as the shared clean baseline leaves the sign unchanged:
\(A=+0.317\) using the image-arm L0 baseline and \(A=+0.203\) using the text-arm L0
baseline. Baseline resampling thus affects the estimated magnitude but not the
direction, and we treat the frontier GSM8K experiment as supporting behavioral
evidence rather than part of the primary CLL analysis.

\begin{table}[t]
\centering
\footnotesize
\setlength{\tabcolsep}{3pt}
\renewcommand{\arraystretch}{1.0}
\begin{tabular}{lcc}
\toprule
Model & Image & Text \\
\midrule
Idefics3-8B         & \shortstack{4008/4656\\(86.1\%)}  & \shortstack{2033/2603\\(78.1\%)} \\
\midrule
Phi-3.5-Vision      & \shortstack{1268/1487\\(85.3\%)}  & \shortstack{820/1099\\(74.6\%)} \\
\midrule
Qwen2-VL-2B         & \shortstack{1404/1934\\(72.6\%)}  & \shortstack{1009/1721\\(58.6\%)} \\
\midrule
Qwen2.5-VL-7B       & \shortstack{2436/3893\\(62.6\%)}  & \shortstack{2695/3915\\(68.8\%)} \\
\midrule
LLaVA-OneVision-7B  & \shortstack{3447/4000\\(86.2\%)}  & \shortstack{2082/2772\\(75.1\%)} \\
\midrule
LLaVA-1.6-7B        & \shortstack{1821/2331\\(78.1\%)}  & \shortstack{719/989\\(72.7\%)} \\
\midrule
Arm-pooled          & \shortstack{14384/18301\\(78.6\%)} & \shortstack{9358/13099\\(71.4\%)} \\
\bottomrule
\end{tabular}
\caption{Agreement between generated source choice and CLL-margin sign
(role-neutral arithmetic runs). Cells give count agreeing\,/\,attributable
observations, with the agreement percentage below. Model rows pool GSM8K
and SVAMP over all four legibility levels; the arm-pooled row also pools
across models.}
\label{tab:cll_generation_agreement}
\end{table}

\section{Legibility-Adjusted Regression Details}
\label{app:legibility_adjustment}

The calibration uses the proportional loss of accuracy in a single channel
\(\ell_{M,c,L}=(\mathrm{Acc}_{M,c,0}-\mathrm{Acc}_{M,c,L})/\mathrm{Acc}_{M,c,0}\),
measured by presenting only the degraded channel at level \(L\) and scoring its
source-supported answer.

\emph{Per-model slopes (primary calibration).} We summarize reallocation at each level
by the median CLL-margin shift toward the source that stays clean,
\[
\begin{aligned}
\widetilde{\Delta}^{(I)}_{M,L}
  &= \operatorname{median}_i\!\left(m^{(I)}_{i,L}-m^{(I)}_{i,L0}\right),\\
\widetilde{\Delta}^{(T)}_{M,L}
  &= \operatorname{median}_i\!\left(m^{(T)}_{i,L0}-m^{(T)}_{i,L}\right),
\end{aligned}
\]
and for each model and arm fit a slope through-origin 
\(\widetilde{\Delta}_{M,c,L}=b_{M,c}\,\ell_{M,c,L}\) over
\(L\in\{\mathrm{L0},\mathrm{L2},\mathrm{L4},\mathrm{L5}\}\) (through the origin because
both quantities are zero at L0). The arm contrast is \(D_M=b_{M,T}-b_{M,I}\); we
bootstrap the matched conflict items (10{,}000 resamples), recompute the median shifts,
and refit both slopes with the accuracy-loss estimates held fixed. Per-model slopes for
the arithmetic benchmarks and for ChartQA-Conflict are in
Tables~\ref{tab:arithmetic_calibrated_slopes} and~\ref{tab:chartqa_calibrated_slopes}.

\emph{Pooled regression (complementary check).} As a complementary item-level analysis
we also fit
\[
\begin{aligned}
\Delta_{i,M,c,L}
={}& \beta_0+\beta_1\ell_{M,c,L}+\beta_2T_c\\
&+\beta_3(\ell_{M,c,L}T_c)+\gamma_M+\varepsilon_{i,M,c,L},
\end{aligned}
\]
where \(T_c=1\) for text degradation and \(\gamma_M\) are model fixed effects. Because
\(\ell\) is shared within a model--channel--level cell, we do not treat item rows as
independent evidence for the slope; primary intervals cluster by model, and clustering by
the 36 model--channel--level cells is a complementary check. With only six model
clusters, these results are evidence for the evaluated model set, not a population-wide
estimate.

\begin{table*}[t]
\centering
\small
\setlength{\tabcolsep}{4pt}
\begin{tabular}{llrrrr}
\toprule
Benchmark & Analysis & \(\beta_3\) & SE & 95\% CI & \(p\) \\
\midrule
GSM8K & Item level, model-clustered
& \(+2.705\) & 0.906 & \([+0.375,+5.034]\) & .031 \\
       & Item level, cell-clustered
& \(+2.705\) & 0.828 & \([+1.082,+4.327]\) & .0011 \\
       & Item level, 1\% winsorized, model-clustered
& \(+2.429\) & 0.666 & \([+0.716,+4.143]\) & .015 \\
       & Cell-median aggregation
& \(+1.779\) & 0.967 & \([-0.191,+3.749]\) & .075 \\
\midrule
SVAMP & Item level, model-clustered
& \(+3.273\) & 0.979 & \([+0.756,+5.790]\) & .020 \\
      & Item level, cell-clustered
& \(+3.273\) & 1.103 & \([+1.111,+5.435]\) & .0030 \\
      & Item level, 1\% winsorized, model-clustered
& \(+3.077\) & 0.829 & \([+0.946,+5.208]\) & .014 \\
      & Cell-median aggregation
& \(+1.982\) & 2.054 & \([-2.202,+6.166]\) & .34 \\
\bottomrule
\end{tabular}
\caption{Task-accuracy-adjusted channel interactions under the role-neutral
arithmetic prompt. Positive \(\beta_3\) means that reallocation per unit of
lost unimodal accuracy is larger under text degradation. The item-level
regression estimates a mean shift; the cell-median analysis is a coarser
sensitivity analysis with a different estimand.}
\label{tab:legibility_adjustment}
\end{table*}

\begin{table*}[t]
\centering
\small
\setlength{\tabcolsep}{5pt}
\begin{tabular}{llrrrr}
\toprule
Benchmark & Model
& Image slope \(b_I\)
& Text slope \(b_T\)
& \(D=b_T-b_I\)
& Bootstrap 95\% CI for \(D\) \\
\midrule
GSM8K
& Qwen2-VL-2B
& \(+0.131\) & \(+0.957\) & \(+0.827\)
& \([+0.748,+0.899]\) \\

& Qwen2.5-VL-7B
& \(+1.872\) & \(+0.880\) & \(-0.993\)
& \([-1.149,-0.785]\) \\

& Idefics3-8B
& \(+0.731\) & \(+4.100\) & \(+3.369\)
& \([+3.010,+3.603]\) \\

& LLaVA-OV-7B
& \(+0.021\) & \(+1.648\) & \(+1.627\)
& \([+1.503,+1.722]\) \\

& LLaVA-1.6-7B
& \(+0.006\) & \(+0.570\) & \(+0.564\)
& \([+0.500,+0.623]\) \\

& Phi-3.5
& \(+0.064\) & \(+1.144\) & \(+1.079\)
& \([+0.989,+1.161]\) \\
\midrule
SVAMP
& Qwen2-VL-2B
& \(+0.485\) & \(+1.755\) & \(+1.270\)
& \([+0.925,+1.531]\) \\

& Qwen2.5-VL-7B
& \(+2.981\) & \(+2.279\) & \(-0.702\)
& \([-1.352,-0.250]\) \\

& Idefics3-8B
& \(+1.649\) & \(+7.349\) & \(+5.699\)
& \([+5.127,+6.727]\) \\

& LLaVA-OV-7B
& \(+0.071\) & \(+3.340\) & \(+3.269\)
& \([+2.923,+3.705]\) \\

& LLaVA-1.6-7B
& \(+0.197\) & \(+1.305\) & \(+1.108\)
& \([+0.901,+1.390]\) \\

& Phi-3.5
& \(+0.018\) & \(+3.312\) & \(+3.294\)
& \([+2.828,+3.799]\) \\
\bottomrule
\end{tabular}
\caption{
Per-model reallocation slopes on the rendered-text arithmetic benchmarks after
calibration by proportional unimodal task-accuracy loss. Slopes are fitted
through the origin over L0, L2, L4, and L5. \(b_I\) is movement toward clean
text per unit of lost image-only accuracy, and \(b_T\) is movement toward the
clean image per unit of lost text-only accuracy. Positive
\(D=b_T-b_I\) indicates greater reallocation under text degradation.
Confidence intervals condition on the observed accuracy losses and use 10,000
bootstrap resamples of the matched conflict items.
}
\label{tab:arithmetic_calibrated_slopes}
\end{table*}

\section{The Reversal Survives Accuracy Calibration}
\label{app:calibrated_slopes_chartqa}

A natural objection to the ChartQA reversal is that degrading the chart may
simply destroy more usable information than degrading the report, so that
stronger reallocation away from the chart reflects a larger information loss
rather than a genuine preference. To control for this, we calibrate each
arm's reallocation by how much the same degradation lowers the model's
\emph{unimodal} accuracy---its accuracy when only that source is available.
The slope \(b_I\) (\(b_T\)) then measures reallocation per unit of lost
chart-only (report-only) accuracy, placing the two directions on equal
footing: a model that reallocates only because a source has become less
informative would show no difference between them.
Table~\ref{tab:chartqa_calibrated_slopes} shows the reversal persists:
\(D=b_T-b_I\) stays negative for all six models with 95\% CIs excluding zero,
so it is not explained by chart degradation destroying more information. All six
models exceed the prespecified \(0.10\) clean-accuracy
floor and are calibrated. 

\begin{table*}[t]
\centering
\small
\setlength{\tabcolsep}{5pt}
\begin{tabular}{lrrrr}
\toprule
Model
& Image slope \(b_I\)
& Text slope \(b_T\)
& \(D=b_T-b_I\)
& Bootstrap 95\% CI for \(D\) \\
\midrule
Qwen2-VL-2B    & \(+2.897\) & \(+1.740\) & \(-1.157\) & \([-1.355,-0.895]\) \\
Qwen2.5-VL-7B  & \(+3.138\) & \(+2.003\) & \(-1.134\) & \([-1.309,-0.784]\) \\
Idefics3-8B    & \(+7.215\) & \(+4.901\) & \(-2.314\) & \([-3.456,-1.379]\) \\
LLaVA-OV-7B    & \(+2.637\) & \(+1.575\) & \(-1.062\) & \([-1.330,-0.731]\) \\
LLaVA-1.6-7B   & \(+1.891\) & \(+1.516\) & \(-0.375\) & \([-0.555,-0.229]\) \\
Phi-3.5-Vision & \(+3.411\) & \(+2.810\) & \(-0.601\) & \([-0.951,-0.216]\) \\
\bottomrule
\end{tabular}
\caption{Calibrated slopes on ChartQA-Conflict. \(b_I\) measures reallocation per
unit of lost chart-only accuracy, and \(b_T\) measures reallocation per unit of
lost report-only accuracy. Slopes are fitted through the origin over L0, L2, L4,
and L5. Negative \(D=b_T-b_I\) indicates stronger reallocation per unit of lost
chart accuracy. All six models exceed the prespecified \(0.10\) clean-accuracy
floor and are calibrated; every \(D\) is negative with a 95\% CI excluding zero.
Intervals are paired-bootstrap 95\% confidence intervals over matched items using
10{,}000 resamples.}
\label{tab:chartqa_calibrated_slopes}
\end{table*}

The random-item-intercept variance reached the boundary at zero because the
response is already an L0-referenced within-item change. We therefore do not
use the resulting mixed-model \(p\)-values; Table~\ref{tab:legibility_adjustment}
reports the cluster-aware fixed-effects fits. A separate survival-based
calibration yields \(\beta_3=+5.55\) (\(p=.013\)) on GSM8K and
\(+7.97\) (\(p=.095\)) on SVAMP.

\section{Chart-versus-Table Control}
\label{app:chart_table}

For each of the 229 conflicts we render the item's official ChartQA source
table~\citep{masry2022chartqa} as a plain table image and substitute it for the
chart in the visual slot, keeping the question, report, counterfactual answer
$a_T$, Source~A/B labels, and the L0/L2/L4/L5 degradation ladder identical. We
render only the rows the chart displays, so the table and chart carry the same
information and neither is systematically easier to read. Table values are taken
directly from the source tables and are never edited toward $a_T$; the table
supports $a_I$ exactly as the chart does. Table~\ref{tab:chart_table} reports the
result.

\begin{table*}[t]
\centering
\small
\setlength{\tabcolsep}{6pt}
\begin{tabular}{lrll r}
\toprule
Model & $A_{\mathrm{chart}}$ & $A_{\mathrm{table}}$ [95\% CI]
& table$-$chart (paired) [95\% CI] & $p_{\mathrm{perm}}$ \\
\midrule
Qwen2-VL-2B        & $-1.221$ & $-0.970$\ $[-1.217,-0.813]$ & $+0.125$\ $[-0.040,+0.369]$ & $0.079$ \\
Qwen2.5-VL-7B      & $-1.362$ & $-1.701$\ $[-1.936,-1.523]$ & $-0.335$\ $[-0.481,-0.108]$ & $<.001$ \\
Idefics3-8B        & $-2.175$ & $-2.003$\ $[-2.688,-1.572]$ & $+0.404$\ $[-0.085,+1.257]$ & $0.107$ \\
LLaVA-OneVision-7B & $-1.156$ & $-0.904$\ $[-1.084,-0.656]$ & $+0.098$\ $[-0.028,+0.342]$ & $0.253$ \\
LLaVA-1.6-7B       & $-0.280$ & $-0.386$\ $[-0.677,-0.166]$ & $-0.079$\ $[-0.329,+0.046]$ & $0.278$ \\
Phi-3.5-Vision     & $-0.963$ & $-0.700$\ $[-0.840,-0.370]$ & $+0.355$\ $[+0.098,+0.655]$ & $0.016$ \\
\bottomrule
\end{tabular}
\caption{Chart-versus-table control on ChartQA-Conflict ($n=229$ per model).
$A_{\mathrm{chart}}$ is the median asymmetry with the original chart (as in
Table~\ref{asymmetry_combined}); $A_{\mathrm{table}}$ is the median asymmetry
when the visual source is the official source table rendered as a plain image.
The last column is the within-item paired difference (table $-$ chart); because
it is a within-item median, it need not equal $A_{\mathrm{table}}-A_{\mathrm{chart}}$
from the two columns. All intervals are 95\% paired-bootstrap CIs (10{,}000
resamples); $p_{\mathrm{perm}}$ is a two-sided paired permutation test. Every
$A_{\mathrm{table}}$ interval excludes zero, so the reversal persists under the
table representation. The chart-versus-table difference is not systematic: four
of six intervals include zero, and the two significant changes (Qwen2.5-VL-7B,
Phi-3.5-Vision) point in opposite directions.}
\label{tab:chart_table}
\end{table*}

\section{CLL Normalization Sensitivity}
\label{app:cll_normalization}

The primary CLL margin normalizes each candidate's summed token
log-probability by its answer-token length. To test whether the arm
asymmetry depends on this choice, we define
\[
m_\alpha(c)
=
\frac{\log p(a_T\mid c)}{|a_T|^\alpha}
-
\frac{\log p(a_I\mid c)}{|a_I|^\alpha},
\]
where each log probability is summed over the candidate's answer tokens.
Thus, \(\alpha=0\) uses the unnormalized summed log-probability,
\(\alpha=0.5\) applies partial normalization, and \(\alpha=1\) gives the
primary mean log-probability per answer token. We recompute the item-level
arm contrast using each version of the margin for role-neutral GSM8K
(\(n=1319\)), role-neutral SVAMP (\(n=300\)), and ChartQA-Conflict
(\(n=229\), after the single audited exclusion).

Table~\ref{tab:cll_normalization} shows that the normalization exponent
rescales the magnitude of the asymmetry but does not change its sign or
significance in any model--benchmark cell: every entry is significant
(Wilcoxon signed-rank, \(p<0.05\); most \(p\ll10^{-10}\)), and within each
cell the estimate moves monotonically toward zero as \(\alpha\) increases
without crossing it. The \emph{direction} of the asymmetry is stable under
normalization but differs by benchmark. On both arithmetic benchmarks the
median asymmetry is positive---indicating stronger reallocation under text
degradation---for all models except Qwen2.5-VL-7B, which is near zero on
GSM8K and weakly negative on SVAMP at every \(\alpha\). On ChartQA-Conflict
the asymmetry is negative for all six models at every \(\alpha\), indicating
stronger reallocation under chart (image) degradation. In no case does the
choice of \(\alpha\) alter these conclusions, confirming that the primary
\(\alpha=1\) margin is representative.

\begin{table}[t]
\centering
\small
\setlength{\tabcolsep}{5pt}
\begin{tabular}{l rrr}
\toprule
Model & \(\alpha{=}0\) & \(0.5\) & \(1\) \\
\midrule
\multicolumn{4}{l}{\emph{GSM8K (neutral)}} \\
Qwen2-VL-2B        & \(+2.648\) & \(+1.723\) & \(+1.107\) \\
Qwen2.5-VL-7B      & \(+0.109\) & \(+0.053\) & \(+0.060\) \\
Idefics3-8B        & \(+4.250\) & \(+4.063\) & \(+3.877\) \\
LLaVA-OneVision-7B & \(+4.969\) & \(+3.183\) & \(+1.950\) \\
LLaVA-1.6-7B       & \(+1.432\) & \(+0.928\) & \(+0.616\) \\
Phi-3.5-Vision     & \(+3.172\) & \(+2.072\) & \(+1.310\) \\
\midrule
\multicolumn{4}{l}{\emph{SVAMP (neutral)}} \\
Qwen2-VL-2B        & \(+3.078\) & \(+2.138\) & \(+1.385\) \\
Qwen2.5-VL-7B      & \(-1.257\) & \(-0.926\) & \(-0.731\) \\
Idefics3-8B        & \(+6.356\) & \(+6.031\) & \(+6.000\) \\
LLaVA-OneVision-7B & \(+7.479\) & \(+4.916\) & \(+3.524\) \\
LLaVA-1.6-7B       & \(+2.763\) & \(+1.794\) & \(+1.234\) \\
Phi-3.5-Vision     & \(+6.965\) & \(+4.806\) & \(+3.548\) \\
\midrule
\multicolumn{4}{l}{\emph{ChartQA-Conflict}} \\
Qwen2-VL-2B        & \(-4.796\) & \(-2.493\) & \(-1.221\) \\
Qwen2.5-VL-7B      & \(-5.191\) & \(-2.737\) & \(-1.362\) \\
Idefics3-8B        & \(-5.160\) & \(-3.469\) & \(-2.175\) \\
LLaVA-OneVision-7B & \(-4.077\) & \(-2.205\) & \(-1.156\) \\
LLaVA-1.6-7B       & \(-1.285\) & \(-0.587\) & \(-0.280\) \\
Phi-3.5-Vision     & \(-3.631\) & \(-1.837\) & \(-0.963\) \\
\bottomrule
\end{tabular}
\caption{Median arm asymmetry under alternative candidate-length
normalizations (\(\alpha\in\{0,0.5,1\}\)) for role-neutral GSM8K, role-neutral
SVAMP, and ChartQA-Conflict. Positive values indicate stronger reallocation
under text degradation; negative values, stronger reallocation under image
(chart) degradation. All entries are significant (Wilcoxon signed-rank,
\(p<0.05\)). Within every cell the sign and significance are preserved across
\(\alpha\); the exponent affects only the estimated magnitude.}
\label{tab:cll_normalization}
\end{table}

\section{ChartQA-Conflict Construction}
\label{app:chartqa_construction}

\paragraph{Question eligibility.}
We screened ChartQA questions for whether the chart-supported answer could be
represented and independently verified in a textual evidence report. We retained
questions with an unambiguous semantic reference to a category, series, and value
(e.g., a named country/year/category or a clearly specified aggregate). We excluded
questions that depended on visual position, color, bar height, or an ambiguous series
mapping (e.g., ``the rightmost upper bar'' or ``the green graph''), because their
answer could not be reliably recovered from the source table alone. Arithmetic, ratio,
and comparison questions were retained only when every referenced value and operation
was unambiguous and reproduced the ChartQA gold answer.

\paragraph{Source items and counterfactual answers.}
We start from 449 ChartQA test questions (430 numeric, 19 yes/no); the frozen
ChartQA-Conflict release keeps the 230 numeric-answer items. For each, the original
ChartQA answer is the chart-supported answer \(a_I\), and we build the report-supported
answer \(a_T\) by perturbing \(a_I\) while preserving its type, unit, sign, scale, and
precision, so that \(a_T\) differs from \(a_I\) only after normalization. Whenever
possible we choose a value that does not appear anywhere in the chart. This keeps
attribution clean---an answer equal to \(a_I\) came from the chart, one equal to \(a_T\)
from the report---and prevents a model from producing \(a_T\) by misreading some other
visible chart value. Rather than fixing a distance bound (e.g., \(\pm X\%\) of \(a_I\)),
we choose each perturbation to stay plausible for the question and chart, and discard
any that collapses to \(a_I\) after rounding, falls outside the expected answer range,
or changes the unit, type, sign, or precision.

\paragraph{Table construction.}
For the table ablation, we render each item's official ChartQA source table as a
plain table image containing the same rows, columns, labels, and values as the
corresponding chart. Tables use regular DejaVu Sans for body cells and DejaVu Sans
Bold for headers. The renderer selects the largest font size between 8 and 18 pixels
that fits the complete table; 208 of the 229 tables use the maximum 18-pixel size.
Each table is rendered on a white RGB canvas with exactly the same pixel dimensions
as its corresponding chart (190 items use \(800\times557\) pixels), thereby holding
the image canvas fixed within item. Headers are lightly shaded, rows alternate
between white and light gray, and cells are separated by one-pixel borders. One
unusually long table is split into two side-by-side panels with repeated headers so
that all source values remain visible.

\paragraph{Review.}
Two of the authors, both senior undergraduates, independently reviewed all 230
items, each confirming that (i) the report entails \(a_T\), (ii) \(a_T\) is a valid
answer to the shared question, and (iii) \(a_T \neq a_I\) after normalization; items
flagged by either reviewer were discussed and adjudicated. Because these reviewers
are authors of this paper and were not blind to the counterfactual design, this check
is not independent of the study (see Limitations).

\paragraph{Prompt template.}
The chart and report receive counterbalanced Source~A/B labels, with the A source shown
first:
\begin{quote}
\small
\texttt{Question: \{question\}}\\[2pt]
\texttt{You are given two conflicting evidence sources. Neither source is privileged.}\\
\texttt{Source \{chart label\} is the attached chart.}\\
\texttt{Source \{report label\} is the textual report below:}\\
\texttt{\{report\}}\\[2pt]
\texttt{Respond with exactly one line `\#\#\#\# <answer>', giving only the answer
value with no explanation.}
\end{quote}

\begin{figure*}[t]
\centering
\begin{tcolorbox}[card, width=\textwidth, top=3pt, bottom=4pt,
  title={\faRobot~\textbf{ChartQA-Conflict item}\, (counterbalanced Source~A/B prompt)}]
\small
\ttfamily Question: What is the poverty rate in California in the year 2019?\\
You are given two conflicting evidence sources. Neither source is privileged.
\normalfont

\vspace{4pt}
\begin{minipage}[t]{0.47\textwidth}\vspace*{0pt}
  \begin{tcolorbox}[srcA, title={\faChartBar~Source A\, — attached chart}]
    \centering\includegraphics[width=\linewidth]{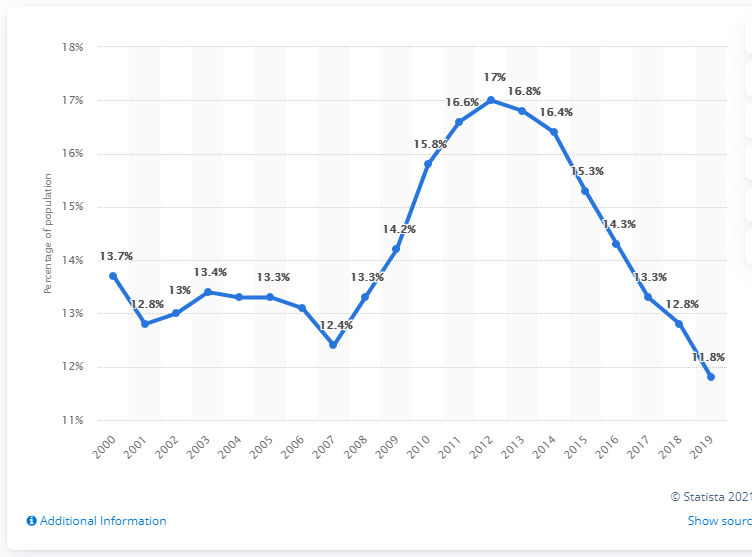}
  \end{tcolorbox}
\end{minipage}\hfill
\begin{minipage}[t]{0.50\textwidth}\vspace*{0pt}
  \begin{tcolorbox}[srcB, title={\faAlignLeft~Source B\, — textual report}]
    \ttfamily\scriptsize The accompanying report summarizes California's poverty
    rate across multiple years, reported as percentages of the population. It
    records a poverty rate of \textbf{14.8\%} in 2019, along with 13.3\% in 2017
    and 14.3\% in 2016.
  \end{tcolorbox}
  \vspace{3pt}
  {\scriptsize
   \faChartBar~Chart (A): 2019 $=$ \textbf{11.8\%} $\Rightarrow a_I{=}11.8$ \emph{(chart-following)}\\[1pt]
   \faAlignLeft~Report (B): 2019 $=$ \textbf{14.8\%} $\Rightarrow a_T{=}14.8$ \emph{(report-following)}}
\end{minipage}

\vspace{3pt}
\ttfamily Respond with exactly one line `\#\#\#\# <answer>'.
\normalfont
\end{tcolorbox}
\caption{A ChartQA-Conflict item as given to the VLM. The question asks for
California's 2019 poverty rate. \textbf{Source~A} is the original chart, whose 2019
value supports $a_I{=}11.8$; \textbf{Source~B} is the evidence-bearing report from
our dataset, perturbed to support $a_T{=}14.8$. A reply of $11.8$ is chart-following,
$14.8$ is report-following; Source~A/B labels are counterbalanced across items. The answer-identification annotations shown below the sources are explanatory overlays and were not included in the model input. Chart
from ChartQA~\citep{masry2022chartqa}.}
\label{fig:chartqa_example}
\end{figure*}

\section{ChartQA-Conflict Behavioral Results}
\label{app:chartqa_behavior}

Generated-answer contrasts (Table~\ref{tab:chartqa_behavior}) use only items that are
decidable at all four required endpoints for a given model, so \(n\) varies and is
smaller than the 229-item CLL sample. The five models with usable generated answers and
CLL scores corroborate the CLL reversal. InternVL2 is behavioral-only because its
custom chat interface does not expose continuation scoring; it shows the
opposite behavioral direction. Phi-3.5-Vision has valid CLL scores and, under the
answer-only extraction, produces well-formed first-line answers, but only nine of its
items are decidable at all four endpoints---too few for a behavioral contrast---so it is
calibrated (Appendix~\ref{app:calibrated_slopes_chartqa}) but not analyzed behaviorally.

\begin{table*}[t]
\centering
\small
\setlength{\tabcolsep}{5pt}
\begin{tabular}{lrrrrrr}
\toprule
Model & \(n\) & \(R_I\) & \(R_T\) & \(A\) & 95\% CI & \(p_W\) \\
\midrule
Qwen2-VL-2B & 113 & \(+.929\) & \(+.053\) & \(-.876\)
& \([-.956,-.788]\) & \(5.63\times10^{-21}\) \\
Qwen2.5-VL-7B & 188 & \(+.957\) & \(+.037\) & \(-.920\)
& \([-.968,-.862]\) & \(1.10\times10^{-36}\) \\
Idefics3-8B & 162 & \(+.722\) & \(+.247\) & \(-.475\)
& \([-.605,-.340]\) & \(7.98\times10^{-10}\) \\
LLaVA-OneVision-7B & 93 & \(+.925\) & \(+.043\) & \(-.882\)
& \([-.968,-.785]\) & \(5.45\times10^{-18}\) \\
LLaVA-1.6-7B & 73 & \(+.808\) & \(+.178\) & \(-.630\)
& \([-.795,-.452]\) & \(5.92\times10^{-8}\) \\
InternVL2-8B & 105 & \(+.286\) & \(+.676\) & \(+.391\)
& \([+.219,+.562]\) & \(4.51\times10^{-5}\) \\
\bottomrule
\end{tabular}
\caption{Generated-answer ChartQA-Conflict contrasts after excluding the item
that failed the report-entailment audit. Values are complete-case mean changes
in source-choice indicators, not CLL margins. Negative \(A=R_T-R_I\) indicates
stronger reallocation under chart degradation. Intervals are paired bootstrap
95\% CIs and \(p_W\) is the two-sided Wilcoxon signed-rank test.}
\label{tab:chartqa_behavior}
\end{table*}

\section{Prompt-Framing Results}
Table~\ref{tab:prompt_framing} compares the original prompt, which presents the
text as the problem and the image as an attachment, with the role-neutral
Source~A/Source~B prompt. The comparison uses the same GSM8K items under both
framings and is available for five models with complete matched CLL results.
Positive values of \(A\) indicate stronger reallocation under text degradation
than under image degradation.

The asymmetry remains positive for every model under both prompts, but framing
changes its magnitude in a model-dependent direction. Neutral framing reduces
the effect for Qwen2.5-VL-7B and Idefics3-8B, increases it for Qwen2-VL-2B and
LLaVA-OneVision-7B, and produces no detectable change for Phi-3.5-Vision.
LLaVA-1.6 is omitted from the CLL comparison because matched original-prompt CLL
results are unavailable; its generated-answer results are available under both
framings.
\label{app:prompt_framing}
\begin{table*}[t]
\centering
\small
\setlength{\tabcolsep}{5pt}
\begin{tabular}{lrrrrr}
\toprule
Model
& \(A_{\mathrm{original}}\)
& \(A_{\mathrm{neutral}}\)
& Neutral--original
& 95\% CI
& Wilcoxon \(p\) \\
\midrule
Qwen2-VL-2B
& \(+0.863\) & \(+1.107\) & \(+0.190\)
& \([+0.125,+0.250]\) & \(9.02\times10^{-16}\) \\

Qwen2.5-VL-7B
& \(+0.813\) & \(+0.060\) & \(-0.375\)
& \([-0.475,-0.290]\) & \(1.90\times10^{-27}\) \\

Idefics3-8B
& \(+4.438\) & \(+3.877\) & \(-0.281\)
& \([-0.625,+0.000]\) & \(2.39\times10^{-4}\) \\

LLaVA-OneVision-7B
& \(+1.596\) & \(+1.950\) & \(+0.215\)
& \([+0.155,+0.259]\) & \(3.56\times10^{-26}\) \\

Phi-3.5-Vision
& \(+1.375\) & \(+1.310\) & \(+0.000\)
& \([-0.085,+0.041]\) & \(0.112\) \\
\bottomrule
\end{tabular}
\caption{Prompt-framing sensitivity on matched GSM8K items. Positive \(A\)
indicates stronger reallocation under text degradation. The
neutral-minus-original contrast is computed within item, so it need not equal
the difference between the two displayed medians. Intervals are paired
bootstrap 95\% confidence intervals, and \(p\) is from a two-sided Wilcoxon
signed-rank test.}
\label{tab:prompt_framing}
\end{table*}

\end{document}